%% file: mfas_arxiv.tex
\documentclass[10pt,twocolumn,letterpaper]{article}
\usepackage{cvpr}
\usepackage{times}
\usepackage{epsfig}
\usepackage{graphicx}
\usepackage{amsmath}
\usepackage{amssymb}

\usepackage[breaklinks=true,bookmarks=false]{hyperref}

\usepackage{multicol}
\usepackage{overpic}
\usepackage{algorithm}
\usepackage{algpseudocode}

\makeatletter
\def\BState{\State\hskip-\ALG@thistlm}
\makeatother

\input{macros}

\cvprfinalcopy 

\def\cvprPaperID{****} 

\ifcvprfinal\fi
\begin{document}

\title{MFAS: Multimodal Fusion Architecture Search}
	
\author{
	\begin{tabular*}{0.666\linewidth}{@{\extracolsep{\fill}}ccc}
		\multicolumn{3}{c}{Juan-Manuel~P\'erez-R\'ua$^{1,3}$\thanks{Assert joint first authorship. This work was done while JMPR was with Orange Labs.} ~~~~~~~~ Valentin~Vielzeuf$^{1,2*}$}
		\\ St\'ephane Pateux$^1$ & Moez~Baccouche$^1$ & Frederic~Jurie$^2$ \\
		\\
		\multicolumn{3}{c}{$^1$Orange Labs, Cesson-S\'evign\'e, France}\\
		\multicolumn{3}{c}{$^2$Universit\'e Caen Normandie, France}\\
		\multicolumn{3}{c}{$^3$Samsung AI Centre, Cambridge, UK}
	\end{tabular*}
}

\maketitle

\begin{abstract}
	We tackle the problem of finding good architectures for multimodal classification problems. We propose a novel and generic search space that spans a large number of possible fusion architectures. In order to find an optimal architecture for a given dataset in the proposed search space, we leverage an efficient sequential model-based exploration approach that is tailored for the problem. We demonstrate the value of posing multimodal fusion as a neural architecture search problem by extensive experimentation on a toy dataset and two other real multimodal datasets. We discover fusion architectures that exhibit state-of-the-art performance for problems with different domain and dataset size, including the \ntu~dataset, the largest multimodal action recognition dataset available.
\end{abstract}

\input{intro.tex}

\input{related.tex}
\input{core.tex}
\input{experiments.tex}
\input{conclusion.tex}

{\small
	\bibliographystyle{ieee}
	\bibliography{biblio}
}

\end{document}

%% file: macros.tex
\newcommand{\ntu}{{\sc Ntu rgb+d}}

\newcommand{\mmimdb}{{\sc Mm-Imdb}}
\newcommand{\avmnist}{{\sc Av-Mnist}}

\newcommand\ignore[1]{}

\def \path{\bp C}

\newcommand{\bff}{{\mathbf{f}}}
\newcommand{\bfg}{{\mathbf{g}}}
\newcommand{\bfh}{{\mathbf{h}}}

\newcommand{\bfx}{{\mathbf{x}}}
\newcommand{\bfy}{{\mathbf{y}}}
\newcommand{\bfz}{{\mathbf{z}}}

\newcommand{\bfW}{{\mathbf{W}}}

%% file: intro.tex
\section{Introduction}
\label{sec:introduction}

Deep neural networks have demonstrated to be effective models for solving a large variety of problems in several domains, including image~\cite{krizhevsky2012imagenet} and video~\cite{baccouche2011sequential} classification, speech recognition~\cite{hinton2012deep}, and machine translation~\cite{wu2016google}, to name a few. 
In a multimodal setting, it is very common to transfer models trained on the individual modalities and merge them at a single point. It can be at the deepest layers, known in the literature as \emph{late fusion}, which is relatively successful on a number of multimodal tasks~\cite{snoek2005early}.
However, fusing modalities at their respective deepest features is not necessarily the most optimal way to solve a given multimodal problem. We argue in this paper that considering features extracted from all the hidden layers of independent modalities could potentially increase performance with respect to only using a single combination of late (or early) features. Thus, this work tackles the problem of finding good ways to combine multimodal features to better exploit the information embedded at different layers in deep learning models for classification.

Our hypothesis is in line with a common interpretation of deep neural models considering that features learned in a convolutional neural network carry varying levels of semantic meanings. In vision, for example, lower layers are known to serve as edge detectors with different orientations and extent, while further layers capture more complex information such as semantic concepts, like \emph{faces, trees, animals}, etc. Evidently, it is difficult to determine by hand what is the most optimal way of mixing features with varying levels of semantic meaning when solving for multimodal classification problems. For example, learning to classify \emph{furry animals} might require analysis of lower level visual features that can be used to build up the concept of fur, whereas classes like \emph{chirping birds} or \emph{growling} might require analysis of more complex audiovisual attributes. Indeed, features from different layers at different modalities can give different insights from the input data. A similar idea is exploited by unimodal ResNets~\cite{he2016deep}, where features from different depths are utilized by later layers through skip connections.

In this line of thought, a few recent works analyzed other possible combinations from input modalities~\cite{shahroudy2017deep, vielzeuf2018centralnet}. 
However, those methods fall short, as the model designer needs to choose empirically which intermediate features to consider.
Evaluating all of the possibilities by hand would be extremely intensive or simply intractable. 
Indeed, the more modalities and the deeper they are, the more complicated it is to choose a mixture. This is all the more true when enabling nested combinations of multimodal features. It is in fact a large combinatorial problem.

In order to handle this issue, the aforementioned combinatorial problem has to be tackled by an efficient search method. Luckily, the underlying structure of this problem makes it specially amenable to sequential search algorithms. We propose in this paper to rely on a sequential model-based optimization (SMBO)~\cite{hutter2011sequential} scheme, which has previously been applied to the related problem of neural architecture search or \emph{AutoML}~\cite{liu2018progressive, epnas2018}. In a few words, we tackle the problem of multimodal classification by directly posing the problem as a combinatorial search. To the best of our knowledge, this is a completely new approach to the multimodal fusion problem, which, as shown by thorough experimentation, improves the state-of-the-art on several multimodal classification datasets. 

This paper brings four main contributions: i) an empirical evidence of the importance of searching for optimal multimodal feature fusion on a synthetic toy database. ii) The definition of a search space adapted to multimodal fusion problems,  which is a superset of modern fusion approaches.	iii) An adaptation of an automatic search approach for accurate fusion of deep modalities on the defined search space. iv) Three automatically-found state-of-the-art fusion architectures for different known and well studied multimodal problems encompassing five types of modalities.

The rest of this paper is organized as follows. In Section~\ref{sec:related}, we describe the work that is related to ours, including multimodal fusion for classification and neural architecture search. Next, in Section~\ref{sec:core} we explain our search space and methodology. In Section~\ref{sec:experiments}, we present an experimental validation of our approach. Finally, in Section~\ref{sec:conclusion}, we give final comments and conclusions.

%% file: related.tex
\section{Related work}
\label{sec:related}

Current design strategies of neural architectures for general classification (multimodal or not) and other learning problems consider the importance of the information encoded at various layers along a deep neural network. 
Indeed, advances in image classification like \emph{residual nets}~\cite{he2016deep} and \emph{densely connected nets}~\cite{huang2017densely} are related to this idea. 
Similarly, for the problem of pose estimation, \emph{stacked hourglass networks}~\cite{newell2016stacked} connect encoder and decoder parts of an autoencoder by short-circuit convolutions, allowing the final classifiers to ponder features from bottom layers. 
However, it is commonly accepted that manually-designed architectures are not necessarily optimally solving the task~\cite{zoph2017neural}. 
In fact, looking at the type of neural networks that are automatically designed by search algorithms, it seems that convoluted architectures with many cross-layer connections and different convolutional operations are preferred~\cite{brock2017smash, zoph2017neural}.

Interestingly, Escorcia~\etal argued that the visual attributes learned by a neural network are distributed across the entire neural network~\cite{escorcia2015relationship}. 
Similarly, it is commonly understood that neural networks encode features in a hierarchical manner, starting from low-level to higher-level features as one goes deeper along them. 
These ideas motivate well our take on the problem of multimodal classification. 
This is, trying to establish an optimal way to connect and fuse multimodal features. To the best of our knowledge, this work is the first one to directly tackle multimodal fusion for classification as an architecture search problem.

In the following, we give an overview of the multimodal fusion problem for classification as a whole. We then continue with a short discussion on relevant methods for architecture search, since it appears at the core of our method.

\paragraph{Multimodal fusion.}

To categorize the different recent approaches of deep multimodal fusion, we can define two main paths of research: architectures and constraints. 

The first path focuses on building best possible fusion \textit{architectures} {\em e.g.} by finding at which depths the unimodal layers should be fused.
Early works distinguished early and late fusion methods~\cite{atrey2010multimodal}, respectively fusing low-level features and prediction-level features. As reported by~\cite{snoek2005early}, late fusion performs slightly better in many cases, but for others, it is largely outperformed by the early fusion. 
Late fusion is often defined by the combination of the final scores of each unimodal branch. This combination can be a simple~\cite{simonyan2014two} or weighted~\cite{natarajan2012multimodal} score average, a bilinear product~\cite{ben2017mutan}, or a more robust one such as rank minimization~\cite{ye2012robust}. Thus, methods such as multiple kernel learning~\cite{bach2004multiple} and super-kernel learning~\cite{wu2004optimal} may be seen as examples of late fusion.
Closer to early fusion, Zhou \etal~\cite{zhou2008feature} propose to use a Multiple Discriminant Analysis on concatenated features, while Neverova \textit{et al}~\cite{neverova2014multi} apply a heuristic consisting of fusing similar modalities earlier than the others. 
Recently, to take advantage of both low-level and high-level features, 
Yang \etal~\cite{yang2016multilayer} leverage boosting for fusion across all layers. To avoid overfitting due to large number of parameters in multilayer approches, multimodal regularization methods~\cite{amer2018deep,gu2017learning,jiang2018exploiting} are also investigated. 
Another architecture approach for multimodal fusion could be grouped under the idea of attention mechanisms, which decides how to ponder different modalities by contextual information.
The mixture of experts by~\cite{jacobs1991adaptive} can be viewed as a first work in this direction. The authors proposed a gated model that picks an expert network for a given input. As an extension, Arevalo \etal~\cite{arevalo2017gated}, proposed Gated Multimodal Units, allowing to apply this fusion strategy anywhere in the model and not only at prediction-level. In the same spirit, multimodal attention can also be integrated to temporal approaches~\cite{hori2017attention,long2018multimodal}.

The second category of multimodal fusion methods proposes to define \textit{constraints} in order to control the relationship between unimodal features and/or the structure of the weights.
Ngiam \etal~\cite{ngiam2011multimodal}, proposed a bimodal autoencoder, forcing the hidden shared representation to be able to reconstruct both modalities, even in the absence of one of them.
Andrew \etal~\cite{andrew2013deep}, adapted Canonical Correlation Analysis to deep neural networks, maximizing correlation between representations. 
Shahroudy \textit{et al.}~\cite{shahroudy2017deep}, use cascading factorization layers to find shared representations between modalities and isolate modality-specific information. 
To ensure similarity between unimodal features, Engilberge \textit{et al.}~\cite{engilberge2018finding} minimize their cosine distance.
Structural constraints can also be applied on the very weights of the neural networks. In addition to modality dropping, Neverova \textit{et al.}~\cite{neverova2016moddrop} propose to zero-mask the cross-modal blocks of the weight matrix in early stages of training. Extending the idea of modality dropping, Li \etal~\cite{li2016modout}, propose to learn a stochastic mask. Another structure constraint as done through tensor factorization was proposed by~\cite{ben2017mutan}.

\paragraph{Neural architecture search.}

The last couple of years have seen an increased interest on \emph{AutoML} methods~\cite{brock2017smash, liu2018progressive, epnas2018, pham2018efficient, zoph2017neural}. Most of these methods rely somehow on a neural module at the core of their respective search approaches. 
This is now known in the literature as \emph{neural architecture search} (NAS). Neural-based or not, \emph{AutoML} methods were traditionally reserved for expensive hardware configurations with hundreds of available GPUs~\cite{liu2018progressive, zoph2017neural}. 

Very recently, progressive exploration approaches and weight-sharing schemes have allowed to tremendously reduce the necessary computing power to effectively perform architecture search on sizeable datasets. Another advantage of progressive search methods \cite{liu2018progressive,epnas2018} is that they leverage the intrinsic structure of the search space, by sequentially increasing the complexity of sampled architectures. In this paper, we start from a sequential method with weight sharing~\cite{epnas2018} and adapt it to the problem of multimodal classification. In particular, we design a search space that is prone to sequential search and which is a superset of previously introduced fusion schemes, \eg,~\cite{vielzeuf2018centralnet}. This is an important aspect of our contribution. As demonstrated by~\cite{zoph2018learning}, constraining the search space is a key element for affordable architecture search. It turns out that directly tackling multimodal datasets by automatic architecture search without designing a constrained, but meaningful, search space would not be tractable.
We demonstrate the value of our approach and the importance of optimizing neural architectures for multimodal classification tasks by tackling three challenging datasets.

%% file: core.tex
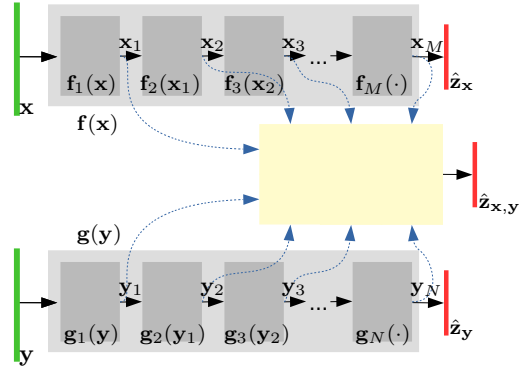
\begin{figure}[t]
	\centering
	\begin{overpic}[width=0.4\textwidth]{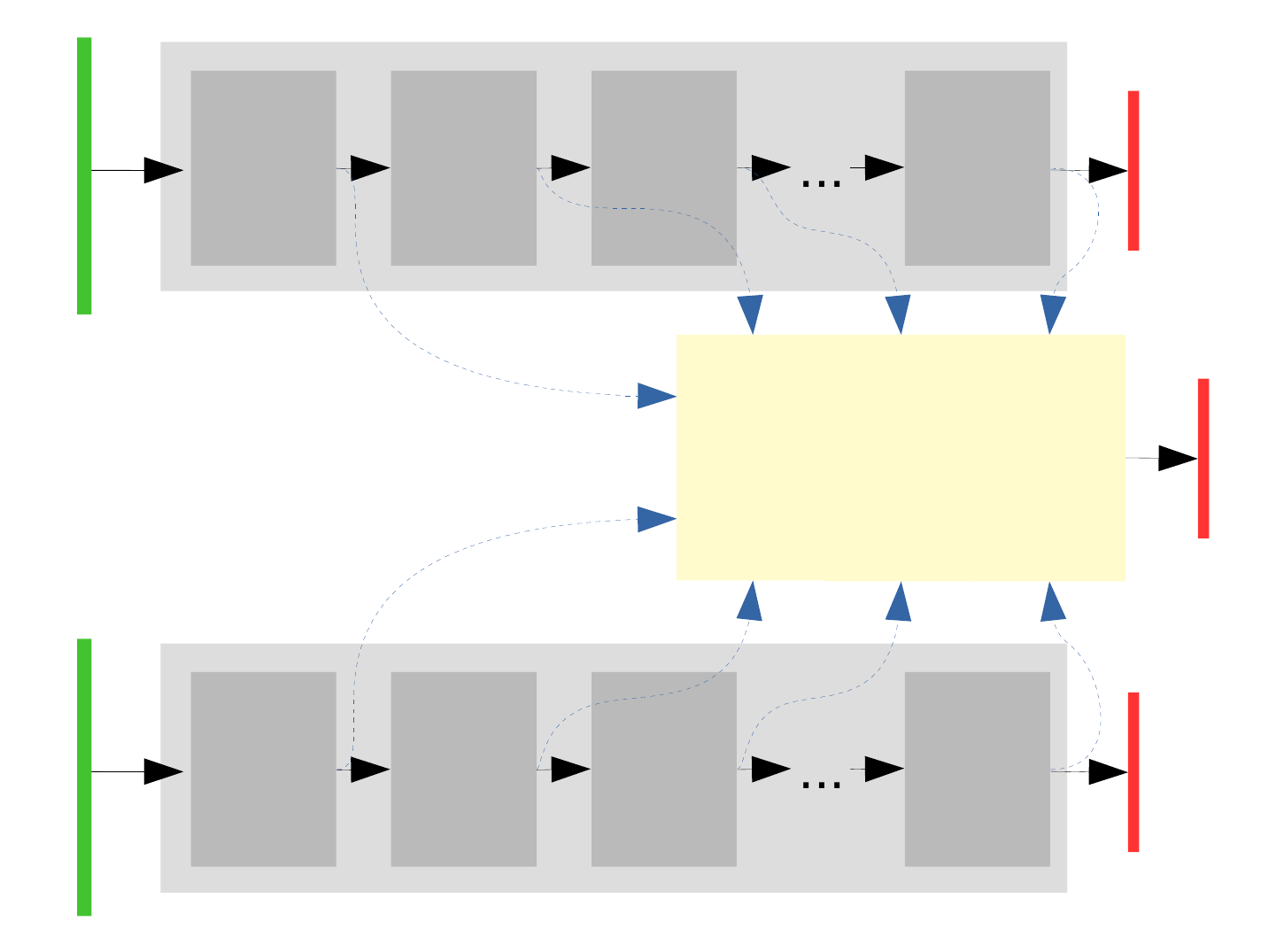}		
		\put (7.0,50) {\small$\displaystyle\bfx$}
		\put (7.0,03) {\small$\displaystyle\bfy$}

		\put (16,54.5) {\footnotesize$\displaystyle\bff_1(\bfx)$}
		\put (15.5,7.0) {\footnotesize$\displaystyle\bfg_1(\bfy)$}		
		\put (26,62.5) {\footnotesize$\displaystyle\bfx_1$}
		\put (26,16.0) {\footnotesize$\displaystyle\bfy_1$}
		\put (18,47) {\footnotesize$\displaystyle\bff(\bfx)$}
		\put (18,26) {\footnotesize$\displaystyle\bfg(\bfy)$}		

		\put (30,54.5) {\footnotesize$\displaystyle\bff_2(\bfx_1)$}
		\put (30,7.0) {\footnotesize$\displaystyle\bfg_2(\bfy_1)$}		
		\put (41.5,62.5) {\footnotesize$\displaystyle\bfx_2$}
		\put (41.5,16.0) {\footnotesize$\displaystyle\bfy_2$}				

		\put (46.0,54.5) {\footnotesize$\displaystyle\bff_3(\bfx_2)$}
		\put (46.0,7.0) {\footnotesize$\displaystyle\bfg_3(\bfy_2)$}					
		\put (57.0,62.5) {\footnotesize$\displaystyle\bfx_3$}
		\put (57.0,16.0) {\footnotesize$\displaystyle\bfy_3$}				

		\put (71.0,54.5) {\footnotesize$\displaystyle\bff_M(\cdot)$}
		\put (71.0,7.0) {\footnotesize$\displaystyle\bfg_N(\cdot)$}
		\put (81.5,62.5) {\footnotesize$\displaystyle\bfx_M$}
		\put (81.5,16.0) {\footnotesize$\displaystyle\bfy_N$}				

		\put (89,55) {\footnotesize$\displaystyle\hat{\bfz}_\bfx$}
		\put (89,08) {\footnotesize$\displaystyle\hat{\bfz}_\bfy$}
		\put (94.5,32) {\footnotesize$\displaystyle\hat{\bfz}_{\bfx,\bfy}$}	
	\end{overpic}
	
	\caption{{\bf General structure of a bi-modal fusion network}. Top: A neural network with several hidden layers (grey boxes) with input $\bfx$, and output $\hat{\bfz}_\bfx$. Bottom: A second network with input $\bfy$, and output $\hat{\bfz}_\bfy$. In this work we focus on finding efficient fusion schemes (yellow box and dotted lines).}
	\label{fig:cnns}
\end{figure}

\section{Methodology}
\label{sec:core}

In this work, as in many others addressing multimodal fusion, we start from the assumption of having an off-the-shelf multi-layer feature extractor for each one of the involved modalities. In practice, this means that we start from a multi-layer neural network for each modality, which we assume to be already pre-trained. However, the reader should consider that our fusion approach is in fact not limited to neural networks as primary feature extractors.

Without loss of conceptual generality, we assume from now on that we will deal with two modalities. The multimodal dataset is composed by pairs of input and output data $(\bfx, \bfy; \bfz)$, where $\bfx$ accounts for the first modality, $\bfy$ for the second one, and $\bfz$ for the supervision labels.
Now, we assume that there exists two functions $\bff(\bfx)$ and $\bfg(\bfy)$ which take $\bfx$ and $\bfy$ as inputs, and output $\hat{\bfz}_\bfx$ and $\hat{\bfz}_\bfy$, which are estimates of the ground-truth labels $\bfz$. 

Furthermore, functions $\bff$ and $\bfg$ are composed of $M$ and $N$ layers, respectively, subfunctions denoted by $\bff_l$ and $\bfg_l$. With a slight abuse of notation, we write for layer $l$, $\bfx_l = (\bff_{l}\circ\bff_{l-1}\cdots\circ\bff_1)(\bfx)$, and $\bfy_l = (\bfg_{l}\circ\bfg_{l-1}\cdots\circ\bfg_1)(\bfy)$. 
See Fig.~\ref{fig:cnns} for a visual representation. Examples of subfunctions when dealing with standard neural networks are operations like convolution, pooling, multiplication by a matrix, non-linearity, etc. The outputs of these subfunctions are the features we want to fuse across modalities. The problem is then to choose which features to fuse and how to mix them. 

\begin{figure*}[t]
	\centering
	\begin{tabular}{cc}
		\begin{overpic}[width=0.50\textwidth]{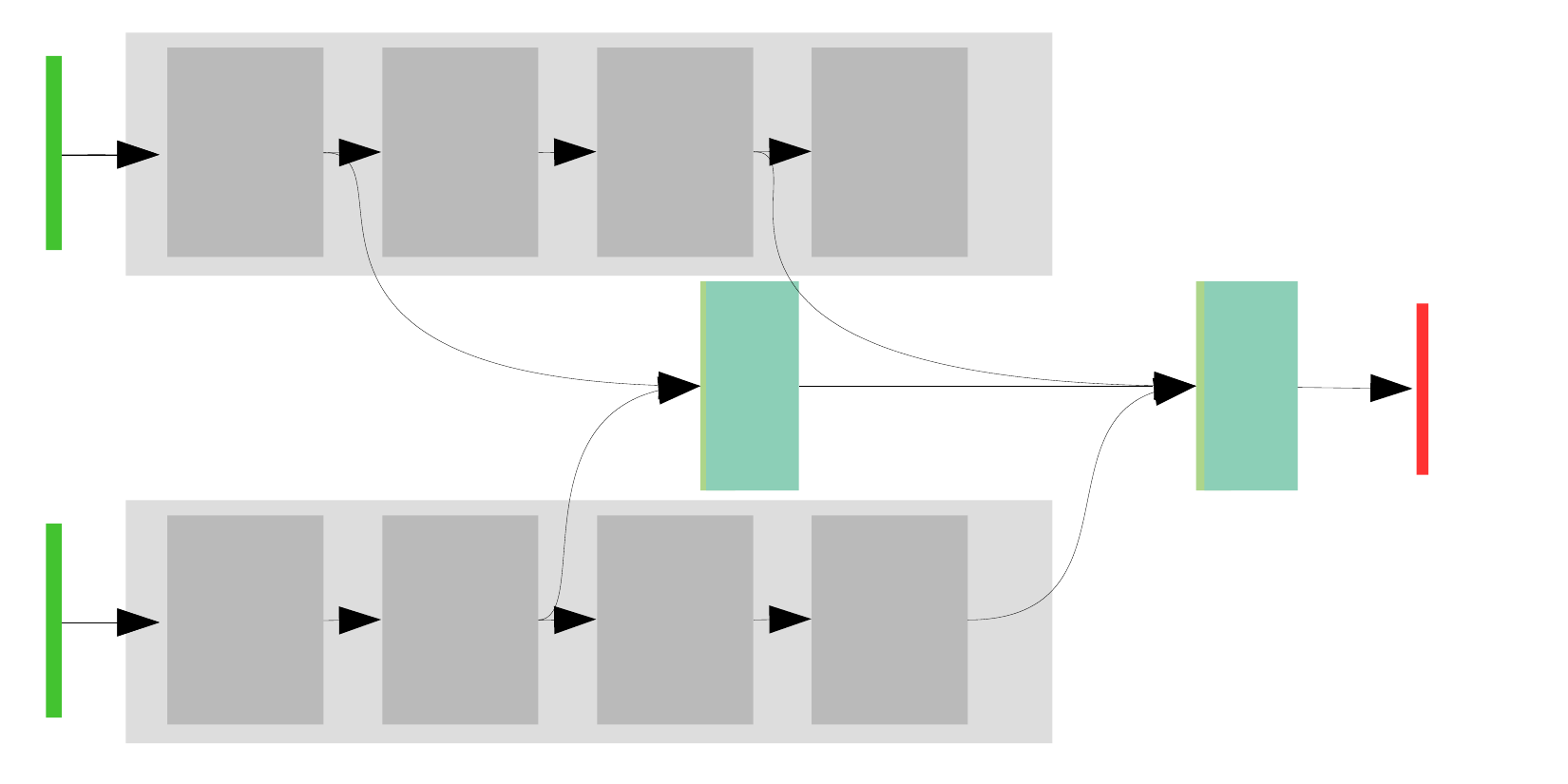}	
			\put (4.0,34) {\small$\displaystyle\bfx$}
			\put (21,42) {\small$\displaystyle\bfx_1$}
			\put (48,42) {\small$\displaystyle\bfx_3$}
			\put (64,33) {\small$\displaystyle\bff$}			
			
			\put (45,19) {\scriptsize$\mathtt{ReLU}$}
			\put (76.5,19) {\scriptsize$\mathtt{Sigm}$}		
			
			\put (4.0,4) {\small$\displaystyle\bfy$}		
			\put (34,7) {\small$\displaystyle\bfy_2$}
			\put (62,7) {\small$\displaystyle\bfy_4$}	
			\put (64,15.5) {\small$\displaystyle\bfg$}			
						
			\put (91.5,19) {\small$\displaystyle\hat{\bfz}_{\bfx,\bfy}$}	
		\end{overpic}
		&
		\begin{overpic}[width=0.50\textwidth]{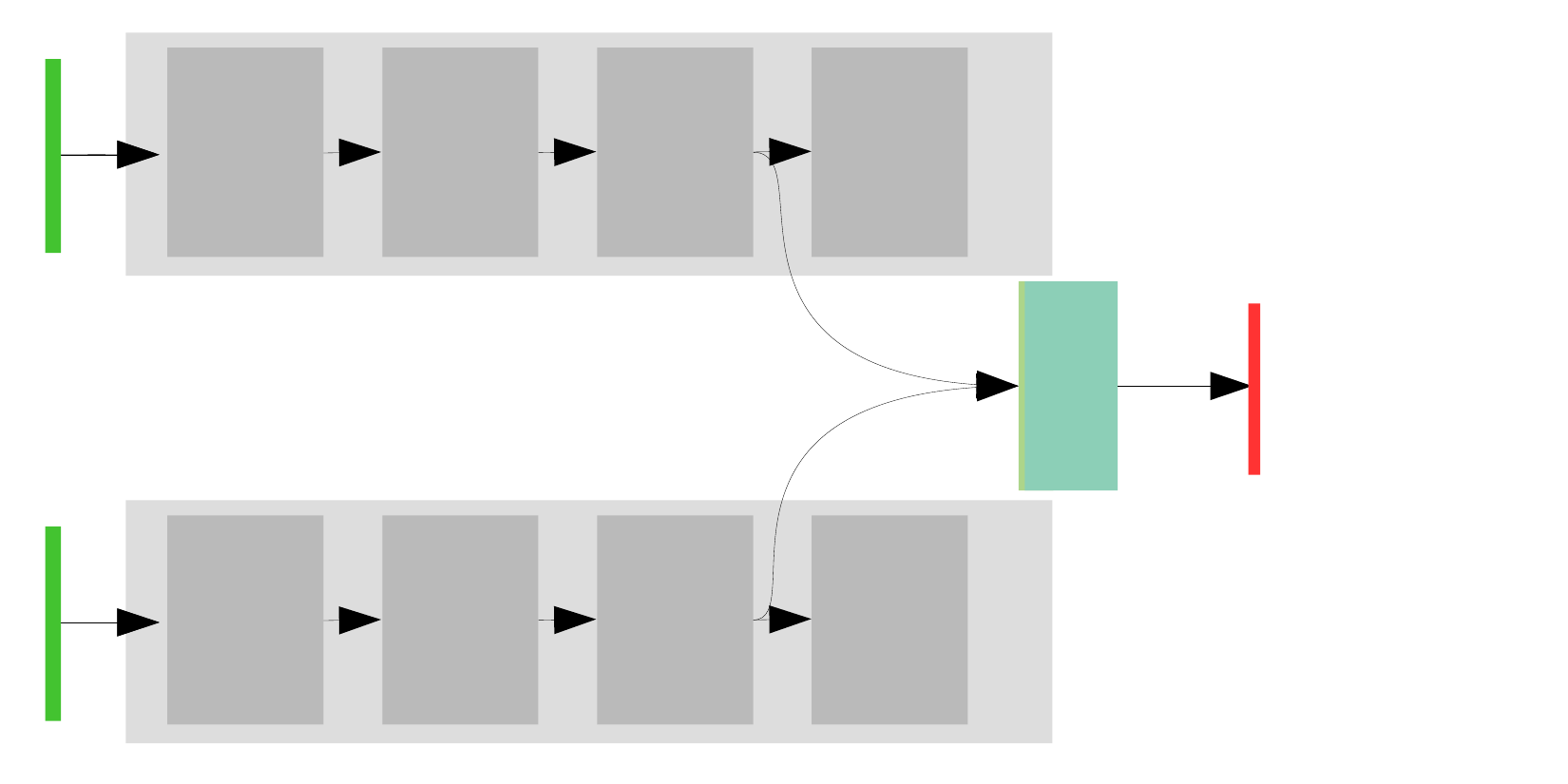}	
			\put (4.0,34) {\small$\displaystyle\bfx$}
			\put (48,42) {\small$\displaystyle\bfx_3$}
			\put (64,33) {\small$\displaystyle\bff$}			
			
			\put (65,19) {\scriptsize$\mathtt{Sigm}$}		
			
			\put (4.0,4) {\small$\displaystyle\bfy$}		
			\put (48,7) {\small$\displaystyle\bfy_3$}
			
			\put (81,19) {\small$\displaystyle\hat{\bfz}_{\bfx,\bfy}$}	
			
			\put (64,15.5) {\small$\displaystyle\bfg$}		
		\end{overpic}	
	\end{tabular}
	
	\caption{{\bf Two realizations of our search space on a small bimodal network}. Left: network defined by $\left[(\gamma^m_1=1, \gamma^n_1=2, \gamma^p_1=1),(\gamma^m_2=3, \gamma^n_2=4, \gamma^p_2=2)\right]$. Right: network defined by $\left[(\gamma^m_1=3, \gamma^n_1=3, \gamma^p_1=2)\right]$.}
	\label{fig:fusion}
\end{figure*}
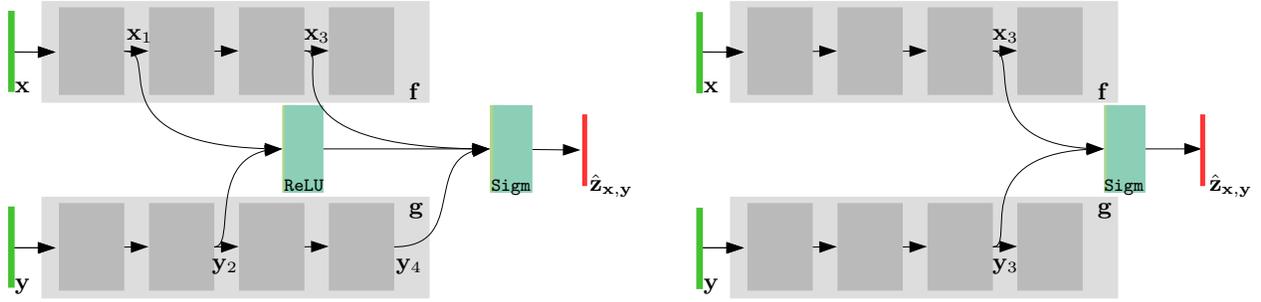

\subsection{Multimodal fusion search space}

In our approach, data fusion is introduced through a third neural network (see Fig.~\ref{fig:fusion} for some illustrations). Each fusion layer $l$ combines three inputs: the output of the previous fusion layer and one output from each modality. This is done according to the following equation:


\begin{equation}
\large
\renewcommand*{\arraystretch}{1.2}
\bfh_l = \boldsymbol{\sigma}_{\gamma^p_l}\left(\bfW_l
\begin{bmatrix}
\bfx_{\gamma^m_l}\\
\bfy_{\gamma^n_l}\\
\bfh_{l-1}      
\end{bmatrix}
\right)
\label{eq:fusion_op}
\end{equation}

\noindent where $\boldsymbol{\gamma}_l=(\gamma^m_l,\gamma^n_l,\gamma^p_l)$ is a triplet of variable indices establishing, respectively, which feature from the first modality, which feature from the second modality, and which non-linearity is applied. 
Also, $\gamma^m_l \in \{1,\cdots,M\}$, $\gamma^n_l \in \{1,\cdots,N\}$, and $\gamma^p_l \in \{1,\cdots,P\}$.  
For the first fusion layer ($l=1$), the fusion operation is defined as:

\begin{equation}
\large
\renewcommand*{\arraystretch}{1.2}
\bfh_1 = \boldsymbol{\sigma}_{\gamma^p_1}\left(\bfW_1
\begin{bmatrix}
\bfx_{\gamma^m_1}\\
\bfy_{\gamma^n_1}
\end{bmatrix}
\right)
\label{eq:fusion_op_l1}
\end{equation}

The number of possible fusion layers, a search parameter, is denoted by $L$, so that $l \in \{1,\cdots,L\}$. 
The fusion layer weight matrix $\bfW_l$ is trainable. Note that we establish feature concatenation as fixed strategy to process and fuse features. In fact, this could be replaced by a weighted sum of input features. However, during our experiments, we noticed that fusion networks with weighted sum of features were almost never chosen, and almost always reduced final classification performance with respect to concatenation. Thus, we decided to simply fix the fusion operation to concatenation.

An illustrative example for $M=N=4$, and $P=2$ ($p=1: \mathtt{ReLU}$; $p=2: \mathtt{Sigmoid}$) is shown in Fig.~\ref{fig:fusion}. We can observe a couple of realizations of the search space for modalities of four hidden layers and two possible non-linearities. On the right, a fusion scheme with a single fusion at the third layer of first and second modalities. On the left, two composed fusions. A composed fusion scheme is defined then by a vector of triplets: $[\boldsymbol{\gamma}_l]_{l\in\{1,\cdots,L\}}$. We denote the set of all possible triplets with $L$ layers as $\boldsymbol{\Gamma}_L$.

Observe that this design enables our space to contain a large number of possible fusion architectures, including the networks defined in, for example, CentralNet~\cite{vielzeuf2018centralnet}. The size of the search space is exponential on the number of fusion layers $L$, and is expressed by: $(M \times N \times P)^L$. 
If we were to tackle a multimodal problem where the number of layers of the feature extractor is only a portion of the depth that modern neural networks exhibit, say $M=N=16$, and only considered two possible non-linearities $P=2$, a fusion scheme with $L=5$ would result in a search space of dimension $\mathbf{\sim3,51\times 10^{13}}$. 

Exhaustively exploring all these possibilities is intractable. In particular, consider that evaluation of a single sample in this space corresponds to training and evaluating a multimodal architecture, which can take from several hours to a few days, depending on the problem at hand. 
This is the reason why we focus on an exploration method that has shown to be sampling-efficient for the related problem of neural architecture search. This is, sequential model-based optimization (SMBO), as used by~\cite{liu2018progressive, epnas2018}. 
In their works, the authors showed that progressively exploring a search space by dividing it into ``complexity levels'', ends up providing architectures that perform as well as the ones discovered by a more direct exploration approach, as in~\cite{zoph2017neural, zoph2018learning}, while sampling fewer architectures. SMBO is well fit to find optimal architectures in the search space designed by~\cite{zoph2018learning}. This is because the space is naturally divided by complexity levels that can be interpreted as progression steps (blocks in the ``micro space''~\cite{liu2018progressive, epnas2018, zoph2018learning}). SMBO sequentially unfolds the complexity of the sampled architectures starting from the simplest one. Luckily, our search space shares a similar structure. We can interpret the number of fusion layers $L$ as a hook for progression.

It is worth noting that the constrained search space that we propose exhibits certain desirable properties. Assuming that the unimodal feature extractor networks are available greatly reduces search burden as they do not need to be trained during search, and the complexity of the problem is limited to a manageable magnitude.

\subsection{Search algorithm}

In SMBO, a model predicting accuracy of sampled architectures lies at the core of the method. This model, or \emph{surrogate} function is trained during progressive exploration of the search space, and it is used to reduce the amount of neural networks that have to be trained and evaluated by predicting performance of unseen architectures. In our case, having a variable-length description of the multimodal architectures $[\boldsymbol{\gamma}_l]_{l\in\{1,\cdots,L\}}$, as described in previous subsection, naturally results in using a recurrent model as \emph{surrogate}. Let us denote this recurrent function by $\boldsymbol{\pi}$. The parameters of $\boldsymbol{\pi}$ are updated at iteration $l$ by stochastic gradient descent (SGD) training on a subset of $\boldsymbol{\Gamma}_l$ with real valued accuracies $\mathcal{A}_l$.

Our procedure, named \emph{multimodal fusion architecture search} (MFAS), and based on ~\cite{liu2018progressive}, is laid out in Alg.~\ref{alg:MFAS}. From lines~\ref{alg:line:ini} to~\ref{alg:line:ine}, the progressive algorithm starts at the smallest fusion network complexity level, \ie, $L=1$. Then, the next complexity levels unroll one after the other by sampling $K$ architectures with a probability that is a function of the surrogate model predictions in lines~\ref{alg:line:compprob} and~\ref{alg:line:sampleK}.
The fusion architecture search is effectively guided by how new architectures are sampled.
Observe that we implement search iterations ($\mathrm{E_{search}}$) and temperature-based sampling ($\mathrm{T_{max}, T_{min}}$) as in EPNAS~\cite{epnas2018}. This is done so the surrogate function does not guide the search with biased assumptions made from partial observations of the search space at early iterations. By using temperature-based sampling, the surrogate function is only trusted as the exploration advances (by reducing the temperature in line \ref{alg:line:updatetemp}). This is complemented by training sampled architectures with very few epochs as in ENAS~\cite{pham2018efficient}, and implementing weight-sharing among sampled architectures to counterweight the main bottleneck of neural architecture search: training sampled architectures to completion. This aspect is of particular importance for multimodal networks, which tend to have a large memory footprint and computing times.

\begin{algorithm}[t]
	\caption{Multimodal fusion architecture search (MFAS)}
	\begin{algorithmic}[1]
		\Procedure{}{{\footnotesize $\bff, \bfg\mathrm{, L, E_{search}, E_{train}, K, S_{train}, S_{val}}, T_{max}, T_{min}$}}
		\footnotesize
		\BState $\mathrm{L}$: max number of fusion layers
		\BState $E_{search}$: number of search iterations
		\BState $E_{train}$: number of training epochs	
		\BState $K$: number of sampled fusion architectures
		\BState $S_{train}, S_{val}$: training and validation sets
		\BState $T_{max}, T_{min}$: sampling temperature range
		
		\State $T \gets \mathrm{T_{max}}$	\emph{\scriptsize // Set temperature}
		\State $\mathcal{B}, \mathcal{A} \gets \lbrace \rbrace$	\emph{\scriptsize // Initialize corresponding sets of arhcs. and accuracy}
		\For{$e = 1 \cdots \mathrm{E_{search}}$} 
			\State $\mathcal{S}_1 \gets \boldsymbol{\Gamma}_1$ \emph{\scriptsize // Set of fusion architectures with $L=1$}   \label{alg:line:ini} 
			\State $\mathcal{M}_1 \gets \mathtt{descToFusionNet}(\mathcal{S}_1, \bff, \bfg)$ \emph{\scriptsize // Build fusion nets} \label{alg:line:buildcnn1}
			\State $\mathcal{C}_1 \gets \mathtt{train}(\mathcal{M}_1, \mathrm{S_{train}}, E_{train})$ \emph{\scriptsize // Train fusion nets}
			\State $\mathcal{A}_1 \gets \mathtt{evaluate}(\mathcal{C}_1, \mathrm{S_{val}})$ \emph{\scriptsize // Get real accuracies for them}	
			\State $\mathcal{B}, \mathcal{A} \gets \mathcal{B} \cup\mathcal{S}_1, \mathcal{A} \cup \mathcal{A}_1$ \emph{\scriptsize // Keep track of sampled archs.}
			\State $\boldsymbol{\pi} \gets \mathtt{update}(\mathcal{S}_1, \mathcal{A}_1)$	\emph{\scriptsize // Train surrogate}   \label{alg:line:ine}
			\For{$l = 2 \cdots L$} 
				\State $\mathcal{S}'_l \gets \mathtt{addLayer}(\mathcal{S}_{l-1}, \boldsymbol{\Gamma}_l)$ \emph{\scriptsize // Unfold 1 more fusion layer}
				\State $\hat{\mathcal{A}'}_l \gets \mathtt{pred}(\mathcal{S}'_l, \boldsymbol{\pi})$ \emph{\scriptsize // Predict with surrogate}	
				\State $\mathcal{P}_l \gets \mathtt{computeProbs}(\hat{\mathcal{A}'}_l, T)$ \emph{\scriptsize // Compute sampling probs.}\label{alg:line:compprob}
				\State $\mathcal{S}_l \gets\mathtt{sampleK}(\mathcal{S}'_l, \mathcal{P}_l, K)$ \emph{\scriptsize // Sample K fusion archs} \label{alg:line:sampleK}	
				\State $\mathcal{M}_l \gets \mathtt{descToFusionNet}(\mathcal{S}_l, \bff, \bfg)$ \emph{\scriptsize // Build fusion net.} \label{alg:line:buildcnn2}
				\State $\mathcal{C}_l \gets \mathtt{train}(\mathcal{M}_l, \mathrm{S_{train}}, E_{train})$ \emph{\scriptsize // Train}	
				\State $\mathcal{A}_l \gets \mathtt{evaluate}(\mathcal{C}_l, \mathrm{S_{val}})$ \emph{\scriptsize // Calculate accuracies}
				\State $\mathcal{B}, \mathcal{A} \gets \mathcal{B} \cup\mathcal{S}_l, \mathcal{A} \cup \mathcal{A}_l$ \emph{\scriptsize // Keep track of sampled archs.}
				\State $\boldsymbol{\pi} \gets \mathtt{update}(\mathcal{S}_l, \mathcal{A}_l)$	\emph{\scriptsize // Update surrogate}
				\State $T \gets \mathtt{updateTemperature}(T, T_{max}, T_{min})$	\label{alg:line:updatetemp}					
			\EndFor
		\EndFor
		
		\Return $\mathtt{topK}(\mathcal{B}, \mathcal{A}, K)$ \emph{\scriptsize // Return best K from all sampled archs.}

		\EndProcedure
	\end{algorithmic}
	\label{alg:MFAS}
\end{algorithm}

Another aspect where our search algorithm differs from the original algorithm~\cite{liu2018progressive} and from~\cite{epnas2018} is that we assume the existence of pre-trained modal functions $\bff$ and $\bfg$. These functions are used to build a multimodal network from a description of the fusion scheme $\mathcal{S}_l$ with $l$ layers (line \ref{alg:line:buildcnn1} and line \ref{alg:line:buildcnn2}). 
At the end of the iterative progressive search, MFAS returns the best $K$ from the set of all sampled architectures $\mathcal{B}$. 

\paragraph{Final architecture.} From Alg.~\ref{alg:MFAS}, we obtain a set of $K$ fusion architectures. One could think of using the surrogate function after its last update to predict the very best fusion scheme from those. However, in this paper we train the best five of the final $K$ architectures to completion, and simply pick the absolute best one from the obtained validation accuracies. During this last training step we also evaluate the performance of the chosen architectures with a larger size of matrices $\bfW_l$. The reduced size is used during search to improve sampling speed and to reduce memory costs.

\paragraph{Loss function.} During the search, the weights of the feature extractors $\bff$ and $\bfg$ are frozen. Because of it, only the fusion softmax $\displaystyle\hat{\bfz}_{x,y}$ is used for the loss function. Found architectures are initially trained for a few epochs with frozen $\bff$ and $\bfg$ functions. A second training step with more epochs involves a multitask loss on $\displaystyle\hat{\bfz}_{x}$, $\displaystyle\hat{\bfz}_{y}$, $\displaystyle\hat{\bfz}_{x,y}$, and unfrozen  $\bff$ and $\bfg$ functions. A categorical cross-entropy loss is used in all the reported experiments unless otherwise noted.

\paragraph{Handling arbitrary tensor dimensions.} A practical issue during the creation of a multimodal neural network from $\bff$ and $\bfg$ is that subfunctions might deliver tensors with arbitrary dimensions, hindering fusion of arbitrary modalities and layer positions. To deal with this in a generic way, we perform global pooling along the channel dimension of 2D and 3D convolutions, while leaving linear layer outputs as they are. 

As a side note, observe in Eq.~\ref{eq:fusion_op} that our default layer type for fusion is fully connected. We experimented with several forms of 1D convolutions without noticing any improvements.

\paragraph{Weight sharing of fusion layers.} In our implementation of Alg.~\ref{alg:MFAS}, multimodal neural networks are not trained in parallel. Instead, sampled fusion networks are trained sequentially for a small number of epochs ($E_{train}=2$ in all of our experiments). For two sample indices $s$ and $s'$, where $s'>s$, we keep track of the weight matrix $\bfW_l^s$ for layer $l$, so $\bfW_l^{s'}$ is initialized from $\bfW_l^s$ if $\mathtt{sizeof}(\bfW_l^s) = \mathtt{sizeof}(\bfW_l^{s'})$. Please note that weights are only shared among matrices in the same layer $l$.

%% file: experiments.tex
\section{Experiments}
\label{sec:experiments}

\begin{table}[tb]
	\caption{Evaluation of our search method on the \avmnist~dataset. The fusion architectures described by arrays of numbers are instances of our search space with $M=3, N=5, P=2$. Validation accuracy is reported.}
	\setlength{\tabcolsep}{6pt}
	\centering
	\small
	\begin{tabular}{c|c|c}
		\hline
		Method & Modalities & Acc \\
		\hline
		\multicolumn{3}{c}{Top-5 found architectures by random search} \\
		\hline
		 $[(3, 3, 2), (5, 3, 2)]$ & image + spect. &  0.9174\\
		 $[(1, 1, 2), (4, 3, 1), (5, 2, 1)]$ & image + spect. &  0.9190\\
		 $[(5, 3, 1), (4, 1, 2)]$ & image + spect. &  0.9196\\		 
		 $[(5, 2, 1), (5, 3, 1)]$ & image + spect. &  0.9224\\		 		 
		 $[(5, 3, 1)]$ & image + spect. &  0.9222\\	
		 \hline
		 Mean (Std) & \multicolumn{2}{c}{0.9203 (0.0021)} \\	 		 		 
		\hline
		\multicolumn{3}{c}{Top-5 found architectures by MFAS} \\		 
		\hline	
		 $[(3, 3, 2), (5, 2, 1), (1, 3, 1), (1, 1, 2)]$ & image + spect. &  0.9258\\
		 $[(5, 2, 1), (5, 2, 2), (5, 1, 1)]$ & image + spect. &  0.9260\\
		 $[(5, 3, 1), (4, 2, 1), (5, 3, 1)]$ & image + spect. &  \textbf{0.9270}\\
		 $[(5, 3, 1), (4, 2, 1), (3, 3, 2)]$ & image + spect. &  0.9266\\		 		 
		 $[(4, 3, 1), (5, 3, 1), (4, 3, 1), (5, 3, 1)]$ & image + spect. &  0.9268\\					
		 \hline
		 Mean (Std) & \multicolumn{2}{c}{0.9264 (0.0004)} \\		 
		\hline
	\end{tabular}
	\label{tab:ss}
	\vspace{-3mm}
\end{table}

In this section we present an extensive experimental validation of our claims. We first start by presenting experiments on a synthetic toy dataset, namely the \avmnist~ dataset~\cite{vielzeuf2018centralnet}. We then continue our experimental work by directly tackling two other multimodal datasets. 
These are i) the visual-textual multilabel movie genre classification dataset by~\cite{arevalo2017gated} (\mmimdb) and ii) the multimodal action recognition dataset by~\cite{shahroudy2016ntu} (\ntu).

For each dataset, we provide a short description of the task as well as the experimental set-up, and then discuss on the results.

\paragraph{\avmnist~dataset.} This is a simple audio-visual dataset artificially assembled from independent visual and audio datasets. The first modality corresponds to $28\times28$ \emph{MNIST} images, with 75\% of their energy removed by PCA. The audio modality is made of audio samples on which we have computed $112\times112$ spectrograms. The audio samples are 25,102 pronounced digits of the \emph{Tidigits} database augmented by adding randomly chosen \textit{noise} samples from the \emph{ESC-50} dataset~\cite{piczak2015esc}. Contaminated audio samples are randomly paired, accordingly with labels, with  MNIST digits in order to reach 55,000 pairs for training and 10,000 pairs for testing. For validation we take 5000 samples from the training set.
The digit energy removal and audio contamination are intentionally done to increase the difficulty of the task (otherwise unimodal networks would achieve almost perfect results and data fusion would not be necessary).

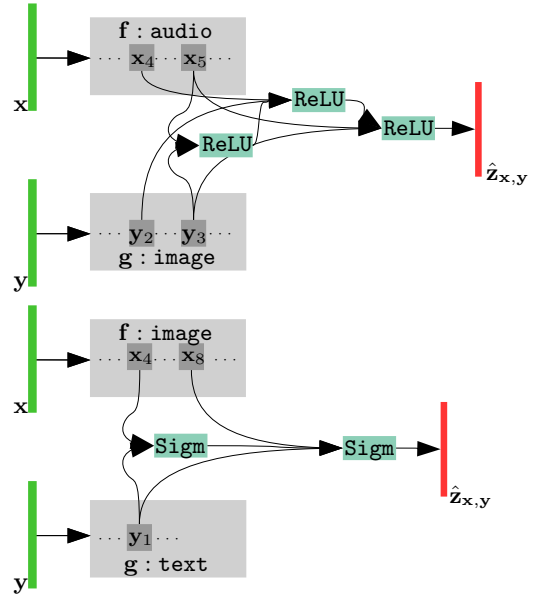
\begin{figure}[t]
	\centering
	\begin{overpic}[width=0.40\textwidth]{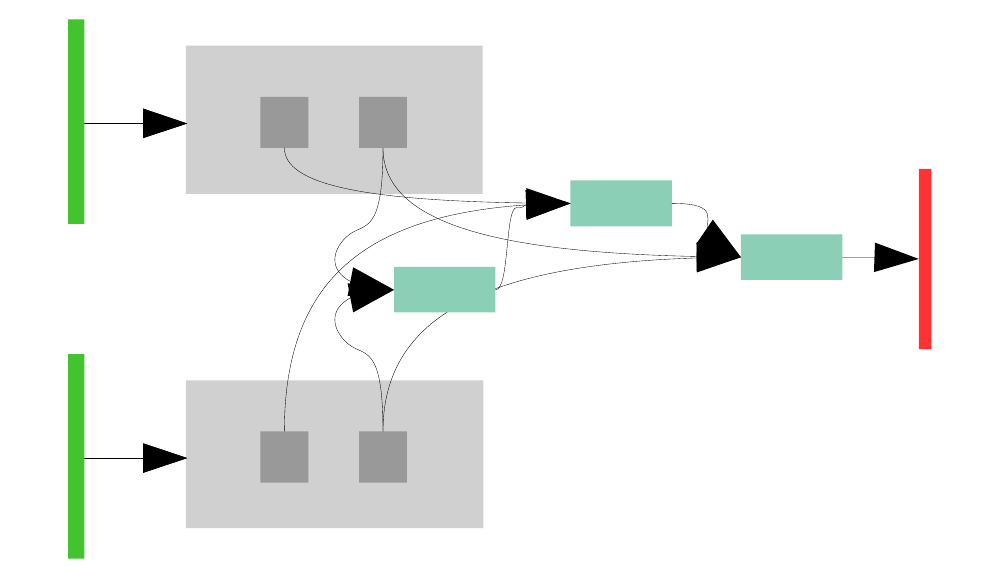}
		\put (4.0,35) {\small$\displaystyle\bfx$}
		\put (26.4,44) {\footnotesize$\displaystyle\bfx_4$}
		\put (36.4,44) {\footnotesize$\displaystyle\bfx_5$}
		\put (24,49) {\small$\displaystyle\bff: \mathtt{audio}$}
		\put (20,44.5) {\tiny$\cdots$}
		\put (30.7,44.5) {\tiny$\cdots$}
		\put (41,44.5) {\tiny$\cdots$}
		
		\put (40.0,27.5) {\small$\mathtt{ReLU}$}
		\put (57.5,36) {\small$\mathtt{ReLU}$}
		\put (74.5,30.5) {\small$\mathtt{ReLU}$}
		
		\put (4,1.5) {\small$\displaystyle\bfy$}	
		\put (26.0,10.5) {\small$\displaystyle\bfy_2$}
		\put (36.0,10.5) {\small$\displaystyle\bfy_3$}
		\put (24,6.0) {\small$\displaystyle\bfg: \mathtt{image}$}
		\put (20,11.0) {\tiny$\cdots$}
		\put (30.7,11.0) {\tiny$\cdots$}
		\put (41,11.0) {\tiny$\cdots$}
		
		\put (94,22) {\small$\displaystyle\hat{\bfz}_{\bfx,\bfy}$}		
	\end{overpic}	
	
	\begin{overpic}[width=0.40\textwidth]{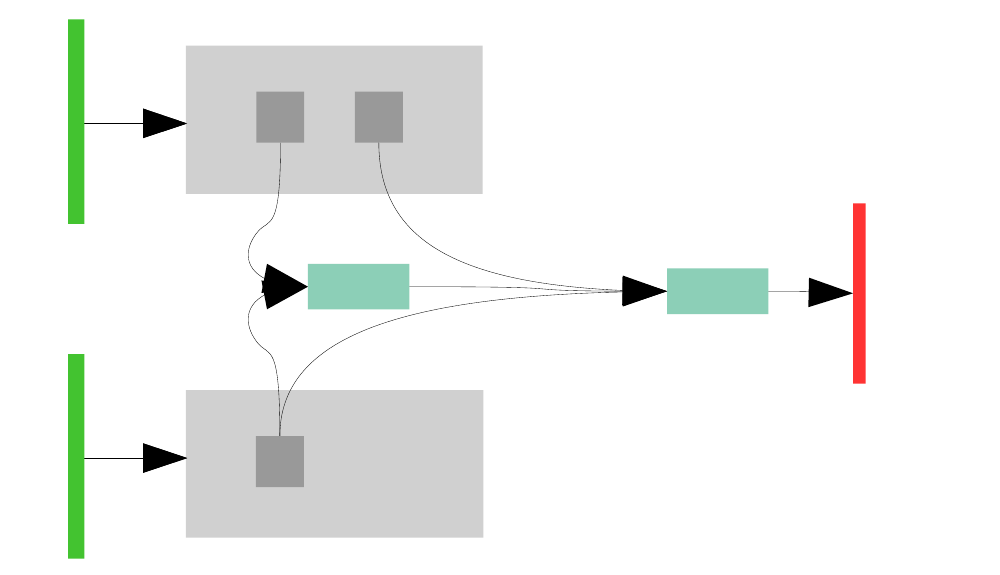}
		\put (4.0,35) {\small$\displaystyle\bfx$}
		\put (20,44.5) {\tiny$\cdots$}
		\put (26,45) {\footnotesize$\displaystyle\bfx_4$}
		\put (30.5,44.5) {\tiny$\cdots$}		
		\put (36,45) {\footnotesize$\displaystyle\bfx_8$}
		\put (24,49) {\small$\displaystyle\bff: \mathtt{image}$}		
		\put (42,44.5) {\tiny$\cdots$}		
				
		\put (31,27.5) {\small$\mathtt{Sigm}$}
		\put (67.0,27.0) {\small$\mathtt{Sigm}$}
		
		\put (4,1.5) {\small$\displaystyle\bfy$}
		\put (20,10.5) {\tiny$\cdots$}
		\put (26,10.5) {\footnotesize$\displaystyle\bfy_1$}
		\put (31,10.5) {\tiny$\cdots$}
		\put (25,5) {\small$\displaystyle\bfg: \mathtt{text}$}
		
		\put (87,18) {\small$\displaystyle\hat{\bfz}_{\bfx,\bfy}$}		
	\end{overpic}	
	
	\caption{{\bf Structure of the found fusion architectures}. First: \avmnist. Second: \mmimdb.}
	\label{fig:found1}
	\vspace{-1mm}
\end{figure}

\begin{table}[tb]
	\caption{Evaluation of multiple fusion architectures on the \avmnist~dataset. Test accuracy is reported.}
	\setlength{\tabcolsep}{6pt}
	\centering
	\small
	\begin{tabular}{c|c|c}
		\hline
		Method & Modalities & Acc (\%) \\
		\hline
		\hline
		\multicolumn{3}{c}{Unimodal baselines for fusion} \\
		\hline		
		LeNet-3~\cite{lecun1990handwritten} & image & 74.52\\	
		LeNet-5~\cite{lecun1990handwritten} & spectrogram & 66.06\\	
		\hline
		\multicolumn{3}{c}{Explicit fusion} \\
		\hline
		Two-stream~\cite{simonyan2014two} & image + spect. & 87.78 \\		
		CentralNet~\cite{vielzeuf2018centralnet} & image + spect. & 87.86 \\
		\hline	
		Ours Top 1 & image + spect. &  \textbf{88.38}\\					
		\hline
	\end{tabular}
	\label{tab:avmnist}
	\vspace{-5mm}
\end{table}

In here, $\bff$ function is a modified LeNet network~\cite{lecun1990handwritten} with five convolutional layers and a global pooling softmax processing spoken digits. Similarly $\bfg$ is a modified LeNet with three convolutional layers. We limit the subfunctions of $\bff$ and $\bfg$ to convolutional layers with $\mathtt{ReLU}$ activation, so we hook global pooling to each one of them: three for the written digit modality ($N=3$), and five for the spectrogram one ($M=5$). For this experiment we let $P=2$ by allowing the activation functions of fusion layers to be either $\mathtt{ReLU}$ or $\mathtt{Sigmoid}$. 

In Table~\ref{tab:ss}, we show results for two exploration approaches: a purely random one (upper part), and MFAS (bottom). Both exploration approaches are allowed to sample 180 architectures. We show validation accuracy for the top five randomly sampled architectures on the proposed search space (top of Table~\ref{tab:ss}). The large standard deviation is a testament to the usefulness of multimodal fusion architecture search. From these results we can infer that some feature combinations provide better insights from data than some other mixtures. At the lower part of Table~\ref{tab:ss} we can see that in contrast to random search, the top five found architectures with our search method present scores with less variability. Furthermore, the best performing architecture on the validation set (in bold) is found by our method.

Test accuracy for baselines and competing fusion architectures are reported in Table~\ref{tab:avmnist}. We report test score of our best found architecture according to Table~\ref{tab:ss}. It can be observed that all multimodal fusion networks largely improve over the unimodal networks, but our automatically found fusion architecture is the one with the best overall score. 
This was found after three iterations of progressive search and $L=4$. The success on this toy (but not trivial) dataset is a first milestone in the validation of our contributions.

\begin{table}[tb]
	\caption{Evaluation of multiple methods on the \mmimdb~ dataset~\cite{shahroudy2016ntu}. Weighted F1 (F1-W) and Macro F1 (F1-M) are reported for each method.}
	\setlength{\tabcolsep}{6pt}
	\centering
	\small
	\begin{tabular}{c|c|c|c}
		\hline
		Method & Modalities & F1-W & F1-M \\
		\hline
		\hline
		\multicolumn{4}{c}{Unimodal baselines for fusion} \\
		\hline		
		Maxout MLP~\cite{goodfellow2013maxout} & text & 0.5754 & 0.4598 \\	
		VGG Transfer & image & 0.4921 & 0.3350 \\	
		\hline
		\multicolumn{4}{c}{Explicit fusion} \\
		\hline
		Two-stream~\cite{simonyan2014two} & image + text & 0.6081 & 0.5049 \\		
		GMU~\cite{arevalo2017gated} & image + text & 0.6170 & 0.5410 \\
		CentralNet~\cite{vielzeuf2018centralnet} & image + text & 0.6223 & 0.5344 \\
		Ours Top 1 & image + text & \textbf{0.6250} & \textbf{0.5568} \\
		\hline
	\end{tabular}
	\label{tab:mmimdb}
	\vspace{-3mm}
\end{table}

\vspace{-2mm}
\paragraph{\mmimdb~dataset.} This multimodal dataset comprises 25,959 movie titles and metadata from the \emph{Internet Movie Database}\footnote{\url{https://www.imdb.com/}}~\cite{arevalo2017gated}. Movie data is formed by their plots, posters (RGB images), genres, and many more metadata fields including director, writer, picture format, etc. The task in this dataset is to predict movie genres from posters and movie descriptions. Since very often a movie is assigned to more than one genre, the classification is multi-label. The loss function used for training is binary cross-entropy with weights to balance the dataset. 

The original split of the dataset is used in our experiments: 15,552 movies are used for training, 7,799 for testing, and 2,608 for validation. The genres to predict include \emph{drama}, \emph{comedy}, \emph{documentary}, \emph{sport}, \emph{western}, \emph{film-noir}, etc., for a total of 23 non-mutually exclusive classes.

Performance of unimodal networks is given at the top of Table~\ref{tab:mmimdb}. Using these unimodal networks as a basis, we implemented Two-stream fusion~\cite{simonyan2014two}, CentralNet~\cite{vielzeuf2018centralnet}, GMU~\cite{arevalo2017gated}, and our best found architecture. One can note that our method gives the best results among the four fusion strategies, once again validating our choices on search space design and fusion scheme~\footnote{Observe  that  the  original  Central-Net paper considers the last features layer (as pre-computed by the original authors~\cite{arevalo2017gated}. Intermediate layers being not provided, we did not start with the exact same unimodal baselines and re-implement all methods in order to allow fair comparison.}. 

The search space for the~\mmimdb~dataset is formed by eight convolutional layers of a VGG-19 image network, and two text Maxout-MLP features. The number of possible fusion configurations available from these features (we set $N=2$, and $M=8$) and the three possible non-linearities ($\mathtt{ReLU}, \mathtt{Sigmoid}, $and $\mathtt{Leaky ReLU}$) is of 110,592. Our best configuration can be seen in Fig.~\ref{fig:found1}.

\paragraph{\ntu~dataset.} This dataset was first introduced by Shahroudy \etal,~\cite{shahroudy2016ntu} in 2016. With 56,880 samples, to the best of our knowledge, it is the largest color and depth multimodal dataset. Capturing 40 subjects from 80 viewpoints performing 60 classes of activities \ntu~is a very challenging dataset with the particularity that it provides dynamic skeleton-based pose data on the top of RGB video sequences. The target activities include \emph{drinking}, \emph{eating}, \emph{falling down}, and even subject interactions like \emph{hugging}, \emph{shaking hands}, \emph{punching}, etc.

\begin{table}[tb]
	\caption{Evaluation of multiple methods on the \ntu~ dataset~\cite{shahroudy2016ntu}. The reported numbers are the average accuracy over the different action subjects (cross-subject measure).}
	\setlength{\tabcolsep}{2pt}
	\centering
	\small
	\begin{tabular}{c|c|c}
		\hline
		Method & Modalities & Acc (\%)\\
		\hline
		\hline
		\multicolumn{3}{c}{Single modality} \\
		\hline
		LSTM \cite{shahroudy2016ntu} & pose & 60.69 \\
		part-LSTM \cite{shahroudy2016ntu} & pose & 62.93 \\
		Spatio-temp. attention \cite{song2017end} & pose & 73.40 \\
		\hline
		\multicolumn{3}{c}{Multiple modalities} \\
		\hline
		Shahroudy \etal~\cite{shahroudy2017deep} & video + pose & 74.86 \\
		Shahroudy \etal~\cite{shahroudy2017deep} & video + pose & 74.86 \\
		Bilinear Learning~\cite{hu2018deep} & video + pose & 83.30 \\		
		Bilinear Learning~\cite{hu2018deep} & video + pose + depth & 85.40 \\
		2D/3D Multitask~\cite{luvizon20182d} & video + pose & 85.50 \\
		\hline
		\multicolumn{3}{c}{Unimodal baselines for fusion} \\
		\hline		
		Inflated ResNet-50~\cite{baradel2018glimpse} & video & 83.91 \\	
		Co-occurrence~\cite{li2018co} & pose & 85.24 \\	
		\hline
		\multicolumn{3}{c}{Explicit fusion} \\
		\hline
		Two-stream~\cite{simonyan2014two} & video + pose & 88.60 \\		
		GMU~\cite{arevalo2017gated} & video + pose & 85.80 \\				
		CentralNet~\cite{vielzeuf2018centralnet} & video + pose & 89.36 \\
		Ours Top 1 & video + pose & \textbf{90.04}$\pm0.6$ \\
		\hline
	\end{tabular}
	\vspace{-3mm}
	\label{tab:ntu}
\end{table}

\begin{table}[ht]
	\centering
	\begin{tabular}{ccc}
		\hline
		Network & \# of fusion Parameters & Acc (\%) \\ \hline
		0       & 2,229,248  & 0.9327           \\
		1       & 2,196,480  & 0.9289           \\
		2       & 1,737,728  & 0.9301           \\
		3       & 2,163,712  & \textbf{0.9346}           \\ \hline
	\end{tabular}
	\caption{Top 4 found architectures on \ntu~according to validation accuracy during search.}
	\vspace{-3mm}
	\label{tab:ntu_found}
\end{table}

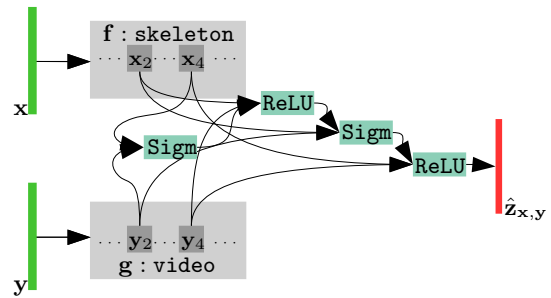
\begin{figure}[t]
	\centering
	\begin{overpic}[width=0.40\textwidth]{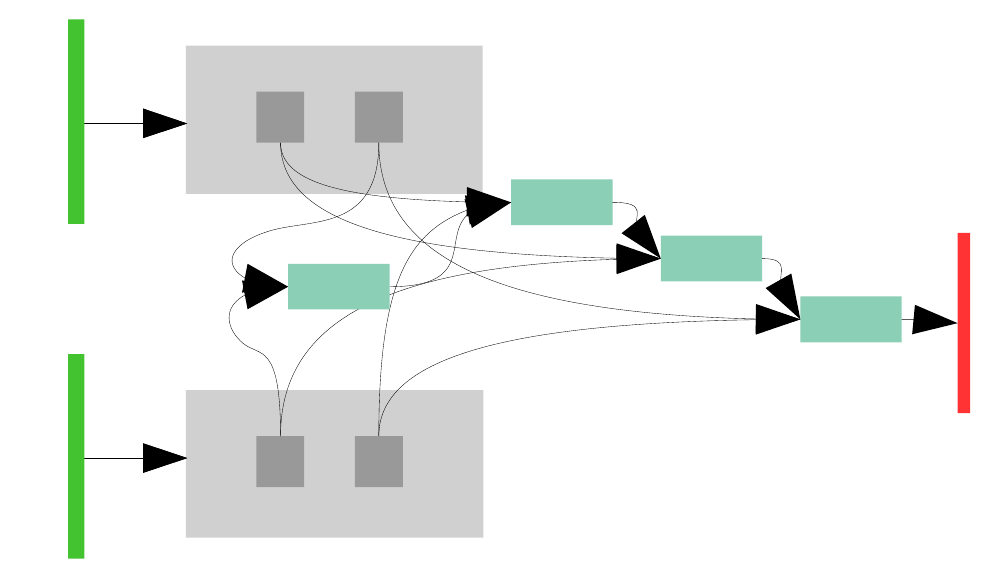}
		\put (4.0,35) {\small$\displaystyle\bfx$}
		\put (26,44.5) {\footnotesize$\displaystyle\bfx_2$}
		\put (36,44.5) {\footnotesize$\displaystyle\bfx_4$}
		\put (20,45.0) {\tiny$\cdots$}
		\put (30.5,45.0) {\tiny$\cdots$}
		\put (42,45.0) {\tiny$\cdots$}		
		\put (21,49) {\small$\displaystyle\bff: \mathtt{skeleton}$}		
		\put (24,5) {\small$\displaystyle\bfg: \mathtt{video}$}
		
		\put (29.5,27.5) {\small$\mathtt{Sigm}$}
		\put (51.5,36) {\small$\mathtt{ReLU}$}
		\put (66.5,30.5) {\small$\mathtt{Sigm}$}
		\put (80.5,24.0) {\small$\mathtt{ReLU}$}		
		
		\put (4,1.5) {\small$\displaystyle\bfy$}	
		\put (26,10.0) {\footnotesize$\displaystyle\bfy_2$}
		\put (36,10.0) {\footnotesize$\displaystyle\bfy_4$}
		\put (20,10.0) {\tiny$\cdots$}
		\put (30.5,10.0) {\tiny$\cdots$}
		\put (42,10.0) {\tiny$\cdots$}
		
		\put (97.5,16) {\small$\displaystyle\hat{\bfz}_{\bfx,\bfy}$}		
	\end{overpic}	
	
	\caption{{\bf Structure of found fusion architectures}. \ntu.}
	\label{fig:found2}
	\vspace{-3mm}
\end{figure}

In our work, we focus on the cross-subject evaluation, splitting the 40 subjects into training, validation, and testing groups.
The subject IDs for training during search are: 1, 4, 8, 13, 15, 17, 19. For validation we use: 2, 5, 9, and, 14. During final training of the found architectures we use the same splitting originally proposed by~\cite{shahroudy2016ntu}.  
We report results on the testing set to objectively compare our found architectures with manually designed fusion strategies from the state-of-the-art.

Results on the \ntu~dataset are summarized in Table~\ref{tab:ntu}. We report accuracy in percentages for several methods. The first group of methods are models processing single modalities as reported by the authors themselves. The second group of results are by methods from the state-of-the-art processing and fusing several modalities (video, pose, and/or depth). Then, we provide the score as computed by us of methods processing single modalities. For video, we tested the \emph{Inflated ResNet-50} used by~\cite{baradel2018glimpse}; and for pose, we leverage the deep co-occurrence model by~\cite{li2018co}. The reported numbers in this group are our departing point and baselines. Finally, the last group of methods perform explicit fusion of modalities and are our main competitors.

Observe that our scores are the highest in Table~\ref{tab:ntu}. We report $90.04\%$ average accuracy over four runs with a variance of $0.6$, which is a significant improvement over all baselines and competing methods. This is achieved by performing fusion search on the convolutional and fully connected features of the Inflated ResNet-50 and deep Co-occurrence baselines. We start from four possible features for each modality ($M=N=4$) and three non-linearities, \ie, $\mathtt{ReLU}, \mathtt{Sigmoid}, $and $\mathtt{Leaky ReLU}$. This means, the search space for the \ntu~dataset is of dimension $5,308,416$. The best found configuration is shown in Fig.~\ref{fig:found2}. In Table~\ref{tab:ntu_found} we report validation accuracy during search for the final top four architectures. Observe that the best architecture is not necessarily the largest one.

\begin{figure}[t]
	\centering
	{\def\arraystretch{0.4}
	\setlength{\tabcolsep}{-0.75em}
	\begin{tabular}{cc}
	\begin{overpic}[width=0.25\textwidth]{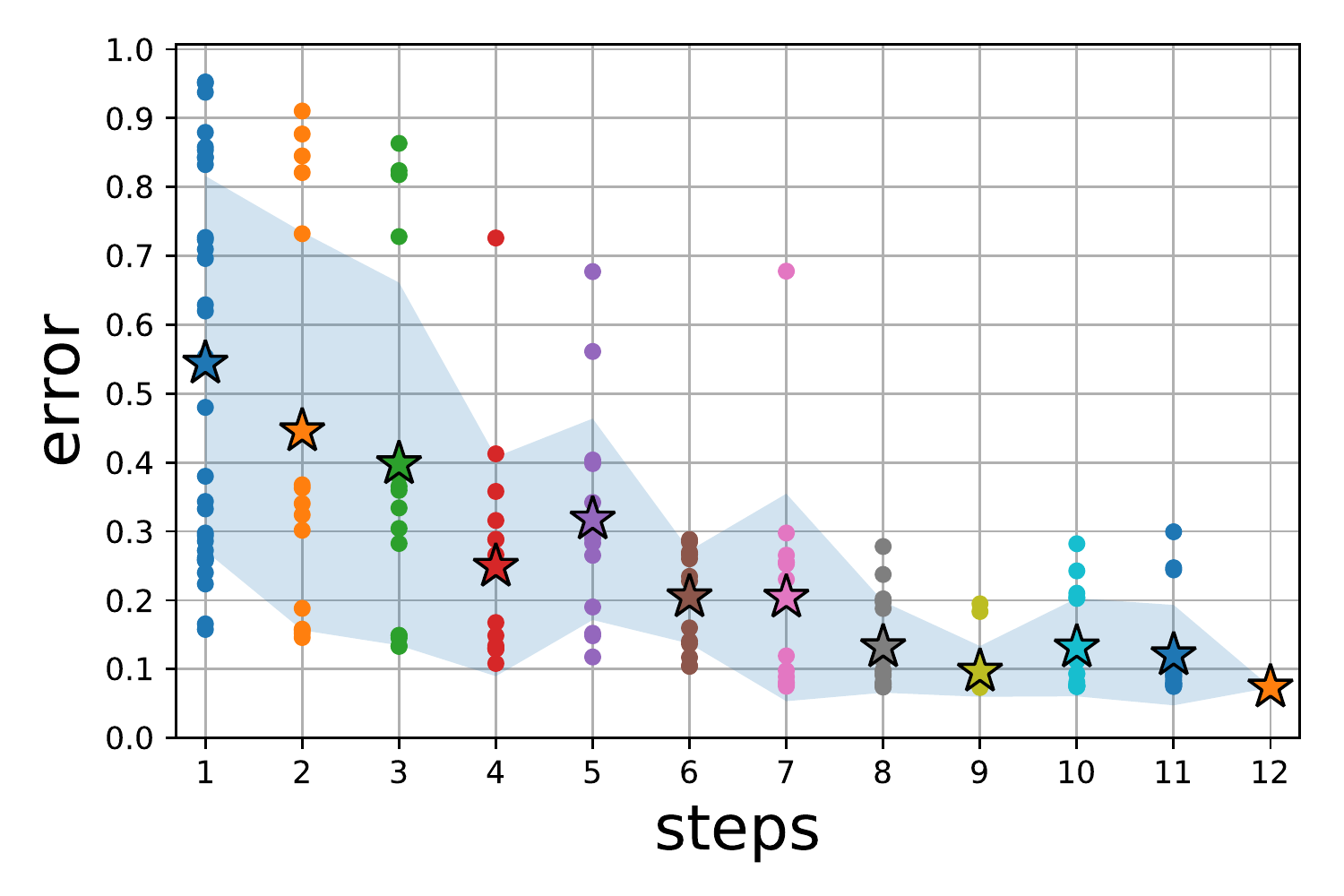}	
	\end{overpic} & 
	\vspace{-2mm}
	
	\begin{overpic}[width=0.25\textwidth]{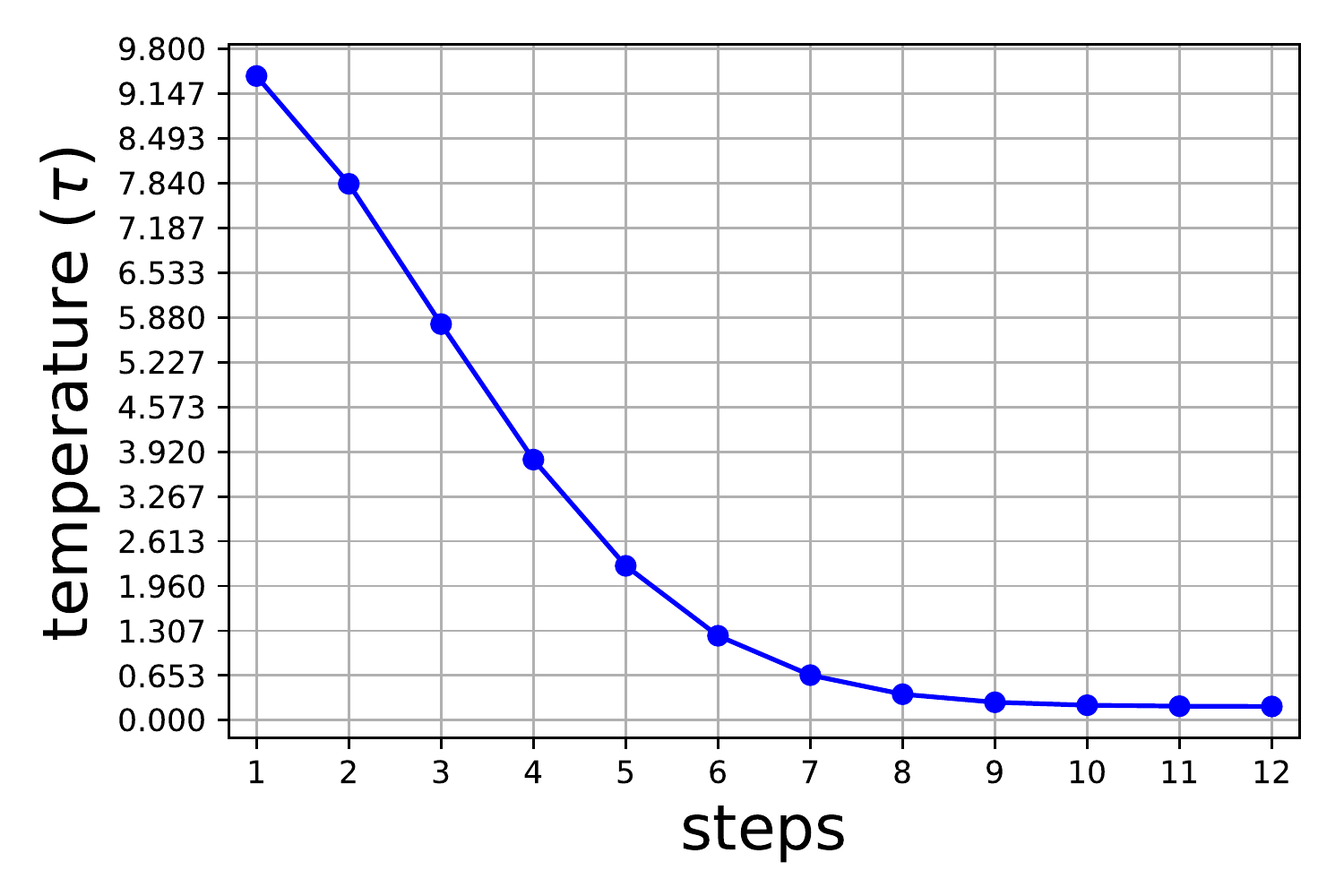}	
	\end{overpic}
	\end{tabular}
	}
	
	\caption{{\bf Left: Error progression during search}. Each plot point represents the validation error of a sampled fusion architecture at a given step of our search algorithm on the \avmnist~set, where the total number of steps is $\mathrm{E_{search}}*L$. Mean error and standard deviation per step are represented with stars and plot shadow, respectively.~{\bf Right: search temperature schedule}.}
	\label{fig:behave}
	\vspace{-5mm}
\end{figure}

\paragraph{Multimodal fusion search behaviour.} In Fig.~\ref{fig:behave} (top) we display the behaviour of our search procedure by plotting validation errors of sampled architectures. It can be observed that, overall, sampled architectures are more and more stable error-wise as the search progresses. The stabilization of sampled errors originates from two sources: first, the shared fusion weights have been more refined at the final steps of the search, and second, the search is driven with more confidence by the predictions of the surrogate function. Indeed, at the last few steps, mean error is significantly lower than the initial ones.

Another interesting effect of our search method and fusion scheme is the fact that even at the initial search steps it is possible to sample architectures that display relatively small validation errors. Since the fusion weights of sampled architectures are trained only for a few epochs, this effect is not necessarily a positive reflection of how good or bad the sampled architecture is. Indeed, it is possible to sample a simple fusion scheme on very deep uni-modal features (which have been pretrained offline) and outperform other sampled architectures that might actually perform better when its weights are revisited at later search steps. In this sense, our temperature-driven sampling of architectures offers a way to escape the fake local minima that originate from this phenomenon. This all boils down to the fact that it is important, in order to avoid getting trapped by initial biased evidence, to trust the surrogate function only after exploration has advanced. We use an inverse exponential schedule for the sampling temperature, as shown at the bottom of Fig.~\ref{fig:behave}, since we observed a better outcome in comparison to a linear temperature schedule.

\paragraph{Search timings.} In Table~\ref{tab:times} we provide the hardware settings and timings for the search on all the reported datasets. Multi GPU training through data parallelism was necessary on the \ntu. Search times on~\ntu~are much larger than on the~\mmimdb~dataset due to model complexity and larger search space. 

\begin{table}[tb]
	\caption{Search timings and hardware configurations.}
	\setlength{\tabcolsep}{3pt}
	\centering
	\small
	\begin{tabular}{c|c|c|c|c}
		\hline
		Dataset & GPUs &  $\mathrm{E_{search}}*L$ & Search time & Avg. step \\
				&(P100)& (steps) & (hours) & time (hours) \\
		\hline
		\hline
		\avmnist& 1 & $3*4 = 12$ & 3.42 & 0.285 \\
		\mmimdb & 1 & $5*3 = 15$ & 9.24 & 0.616 \\
		\ntu    & 4 & $3*4 = 12$ & 150.91 & 12.57\\
		\hline
	\end{tabular}
	\label{tab:times}
	\vspace{-4mm}
\end{table}

%% file: conclusion.tex
\section{Conclusion}
\label{sec:conclusion}

This work tackles the problem of finding accurate fusion architectures for multimodal classification. We propose a novel multimodal search space and exploration algorithm to solve the task in an efficient yet effective manner. The proposed search space is constrained in such a way that it allows convoluted architectures to take place while also containing the complexity of the problem to reasonable levels. We experimentally demonstrated on three datasets the validity of our method, discovering several fusion schemes that provide state-of-the-art results on those datasets. Future research directions include improving the search space so the composition of fusion layers is even more flexible.